%% file: main.tex
\documentclass[journal,twoside,web]{ieeecolor}
\usepackage{tmi}
\usepackage{cite}
\usepackage{amsmath,amssymb,amsfonts}
\usepackage{algorithmic}
\usepackage{graphicx}
\usepackage{textcomp}

\usepackage{bm}
\usepackage{booktabs}
\usepackage{multirow}
\usepackage{balance}

\usepackage{multirow}
\usepackage{hyperref}
\usepackage{footnote}
\usepackage{soul}

\usepackage{everypage}
\usepackage{tikz}
\usepackage{lipsum}% just to generate filler text

\AddEverypageHook{%
 \begin{tikzpicture}[remember picture,overlay, scale=0.6]
 \node[
    scale=0.63,
    opacity=1,
    xshift=0,
    yshift=0,
    color=black
    ]
    at (14, 3.5)  
    {\parbox{24cm}{%
    \begin{tabular}{c}
      This article has been accepted for publication in IEEE Transactions on Medical Imaging. This is the author's version which has not been fully edited \\and content may change prior to final publication. Citation information: DOI 10.1109/TMI.2021.3077285
    \end{tabular}}};
\end{tikzpicture}
}

\AddEverypageHook{%
    \begin{tikzpicture}[remember picture,overlay, scale=0.6]
        \node[
            scale=0.63,
            opacity=1,
            xshift=0,
            yshift=0,
            color=black
        ]
        at (13.5, -41.9)  
        {\copyright 2021 IEEE. Personal use is permitted, but republication/redistribution requires IEEE permisions. See https://www.ieee.org/publications/rights/index.html for more information.};
    \end{tikzpicture}
}

\def\BibTeX{{\rm B\kern-.05em{\sc i\kern-.025em b}\kern-.08em
    T\kern-.1667em\lower.7ex\hbox{E}\kern-.125emX}}
\begin{document}

\title{Object-Guided Instance Segmentation With Auxiliary Feature Refinement for Biological Images}

\author{Jingru Yi, Pengxiang Wu, Hui Tang, Bo Liu, Qiaoying Huang, Hui Qu, Lianyi Han, Wei Fan, \\Daniel J. Hoeppner, Dimitris N. Metaxas 
\thanks{Manuscript received April 30, 2021.}
\thanks{J. Yi, P. Wu, B. Liu, Q. Huang, H. Qu, D. N. Metaxas are with the Department of Computer Science, Rutgers University, Piscataway, NJ 08854, USA (e-mails: jy486@cs.rutgers,edu, pw241@cs.rutgers.edu, kfliubo@gmail.com, qh55@cs.rutgers.edu, hq43@cs.rutgers.edu, dnm@cs.rutgers.edu).}
\thanks{H. Tang, L. Han and W. Fan are with Tencent Hippocrates Research Labs, Palo Alto, CA 94306, USA (e-mails: tanghui@tencent.com, lianyihan@tencent.com, davidwfan@tencent.com).}
\thanks{D. J. Hoeppner is with Lieber Institute for Brain Development, MD 21205, USA (e-mail: daniel.hoeppner@astellas.com).}
}

\maketitle

\begin{abstract}
Instance segmentation is of great importance for many biological applications, such as study of neural cell interactions, plant phenotyping, and quantitatively measuring how cells react to drug treatment. In this paper, we propose a novel box-based instance segmentation method. Box-based instance segmentation methods capture objects via bounding boxes and then perform individual segmentation within each bounding box region. However, existing methods can hardly differentiate the target from its neighboring objects within the same bounding box region due to their similar textures and low-contrast boundaries. To deal with this problem, in this paper, we propose an object-guided instance segmentation method. Our method first detects the center points of the objects, from which the bounding box parameters are then predicted. To perform segmentation, an object-guided coarse-to-fine segmentation branch is built along with the detection branch. The segmentation branch reuses the object features as guidance to separate target object from the neighboring ones within the same %cropped 
bounding box region.  To further improve the segmentation quality, we design an auxiliary feature refinement module that densely samples and refines point-wise features in the boundary regions.
Experimental results on three biological image datasets demonstrate the advantages of our method. The code will be available at \url{https://github.com/yijingru/ObjGuided-Instance-Segmentation}.
\end{abstract}

\begin{IEEEkeywords}
Instance segmentation, medical image detection, medical image segmentation
\end{IEEEkeywords}

\section{Introduction}
\label{sec:introduction}
\IEEEPARstart{I}{nstance} segmentation is a prerequisite for many biological applications such as phenotype measurement of cells \cite{boutros2015microscopy} and plants \cite{minervini2016finely,kumar2019image}, cell lineage study \cite{ravin2008potency} and  histopathological image analysis \cite{xie2019deep,zhou2019cia,li2019accurate}. It is a challenging task since it is non-trivial to separate the attached objects when their boundaries are of low contrast (e.g., overlapped leaves), as well as to capture the fine-grained object details (e.g., the filopodia and lamellipodia of neural cells). Instance segmentation requires assigning both class label (i.e., foreground and background) and instance label (i.e., object identity) to each pixel of an input image. The challenges of this task involve instance clustering, occlusion, adhesion of objects, etc. Existing instance segmentation methods can be divided into two types: box-free and box-based instance segmentation.

Box-free instance segmentation methods capture instances by analyzing the object morphology properties (e.g., boundaries, shapes, and textures) without the aid of object bounding boxes. In traditional methods this is typically achieved by extracting hand-crafted features. For example, Quelhas et al. \cite{5477157} use a carefully designed sliding band filter to detect and segment cell nuclei. Tareef et al. \cite{8314741} propose a fast watershed method to capture cervical cells. Bamford et al. \cite{bamford1998unsupervised} adopt active contours for cell nucleus segmentation. Boomgaard et al. \cite{nguyen2003watersnakes} propose an energy-driven watershed method to improve the segmentation.
These methods are well motivated with theoretical insights, but are sensitive to intensity variations and suffer from under- or over-segmentation. With the advancement of deep learning methods in computer vision community, in recent years researchers have developed more robust methods with the aid of convolutional neural networks (CNN). One representative example is DCAN \cite{chen2017dcan}, which adopts fully convolutional network (FCN) to predict semantic labels of pixels and separates the touching objects by fusing contour information. Unlike DCAN, RecurrentIS \cite{romera2016recurrent} uses convolutional LSTM \cite{sonderby2015convolutional} to recurrently separate the leaf instances from FCN feature maps. Deep Watershed \cite{bai2017deep} learns an energy landscape and extracts segmentation masks according to the energy surfaces. StarDist \cite{schmidt2018cell} learns star-convex polygons to describe cell shapes. CosineEmbedding \cite{payer2018instance} retrieves instance masks by clustering pixel embeddings. Li et al. \cite{li2019accurate} differentiate the nuclear masks through their center masks. CIA-Net \cite{zhou2019cia} aggregates the multi-level contour information to separate adjacent nuclei instances. Yan et al. \cite{yan2020enabling} measure the segment-level shape similarities to differentiate gland instances, a method which is limited to objects with similar shapes. PatchPerPix \cite{mais2020patchperpix} assembles connected instances from predicted shape descriptors. CellPose \cite{stringer2020cellpose} performs instance segmentation based on a new dataset comprising highly varied images of cells.

% input figure 1
\input{images/fig1}

Box-based instance segmentation comprises both object detection and object segmentation. In these methods, an object detector (e.g.,\cite{girshick2015fast,NIPS2015_5638,girshick2014rich,liu2016ssd}) first locates objects using bounding boxes; then instance segmentation is performed locally within regions of interest (ROI). 
One representative work is Mask R-CNN \cite{he2017mask}, which adopts faster R-CNN \cite{NIPS2015_5638} to detect objects. It builds a segmentation head along with the detection head to output bounding boxes and segmentation masks simultaneously. Similarly, Ren et al. \cite{ren2017end} provide a box net to localize and segment the instances recursively using attentive recurrent network. MaskLab \cite{Chen_2018_CVPR} employs faster R-CNN and direction pooling for instance identification. To enhance the performance of instance segmentation, a few works focus on improving the object detector, which is crucial for the success of box-based strategy. For example, BAIS \cite{hayder2017boundary} corrects the bounding box shift issues with boundary-aware masks. PANet \cite{liu2018path} aggregates multi-layers' information to boost the detection accuracy of feature pyramid network (FPN) \cite{lin2017feature}. ANCIS \cite{YI2019228} employs attentive single shot multibox detector (SSD) \cite{liu2016ssd} to locate and segment cells in a coarse-to-fine manner. These methods have demonstrated promising results; however, the anchor-based object detectors used in them have several disadvantages. First, the anchor boxes are assigned to multiple feature maps with different scales and aspect ratios. However, only a small number of anchor boxes overlap with the ground-truth objects. This leads to an imbalance issue between positive and negative anchor boxes, making it difficult to train the object detector. %This 
Second, the anchor boxes introduce more hyper-parameters and design choices, which are non-trivial to select in practice. Recently, keypoint-based object detectors (e.g.,\cite{law2018cornernet,zhou2019bottom,duan2019centernet}) are developed to address the above issues, and have shown better performances than anchor-based detection methods. Such a detection strategy has been incorporated into instance segmentation methods. For example, Keypoint Graph \cite{YiMICCAI2019} detects the center and four corner points of a bounding box. Afterwards, the five points are grouped to form a predicted bounding box for the target object. Although being accurate for large objects, Keypoint Graph typically fails to capture small objects due to the overlapping of keypoints. Besides, the post-grouping process is computationally expensive. 

In this paper, we propose to localize objects by detecting their center points, from which the bounding box parameters are determined. Different from Keypoint Graph \cite{YiMICCAI2019} that groups the five points to form a bounding box, our method infers the bounding boxes directly from the center points. Without post-grouping process, the proposed method is computationally efficient. Moreover, only detecting the center points facilitates the identification of small objects since it does not have the overlapping issue as in Keypoint Graph. In addition, compared to our previous work, i.e., ANCIS \cite{YI2019228}, the proposed method is anchor-box free. %Without s
Apart from object detector, a good segmentation method also plays an important role in instance segmentation task. To predict instance mask for each cropped ROI region, a few works construct a separate coarse-to-fine segmentation branch along with the object detection branch; examples include\cite{ YiMICCAI2019,YI2019228,ronneberger2015u}. The coarse-to-fine segmentation branch utilizes multi-level ROI regions cropped from the backbone network and formulates a U-shaped \cite{ronneberger2015u} network that performs full-resolution segmentation for each instance. However,  the U-shaped coarse-to-fine architecture can hardly differentiate the target object from the neighboring instances within the same ROI region due to their similar textures and blurred boundaries. To deal with this weakness, %Arnab 
in this paper, we build an object-guided segmentation branch that reuses object features %as a guide
to make the model learn to remove undesired neighboring objects within a cropped ROI region.

To further improve the segmentation quality, we design an auxiliary feature refinement module. Our motivation is as follows. In the segmentation task, the network assigns class labels (i.e., foreground and background) to pixels sampled from regular grids. In this process, the model tends to sample more points from smooth regions (i.e., the areas inside object boundaries), while high-frequency regions (i.e., the areas around object boundaries) are under-sampled. As a result, during training the model is biased towards the smooth regions yet features in the high-frequency regions are not sufficiently learned. To address this issue, DCAN \cite{chen2017dcan} makes the network predict a contour map to explicitly separate connected objects. A similar strategy is to use a pre-computed weight map to make the network focus on separation borders \cite{ronneberger2015u}. Another strategy is to employ subdivision \cite{whitted2005improved} or adaptive sampling \cite{mitchell1987generating} to refine the sampling point features in the boundary regions. For instance, PointRend \cite{kirillov2019pointrend} formulates image segmentation as a rendering problem and computes high-resolution segmentation maps by densely sampling and labeling the boundary pixels. However, PointRend would lead to non-smooth segmentation due to inconsistent interpolation in the output segmentation mask. Besides, PointRend fixes a very large resolution (e.g., $224\times224$) for each ROI, but in practice a size-adaptive scheme is more natural and cost-effective. In this paper, we propose an auxiliary feature refinement module that densely samples and refines the boundary features only in the training process. In this way, we make the model focus more on the boundary regions during training and retain a consistent pixel labeling for the segmentation output.

\subsection{Contributions}
\label{sec:contributions}
In this paper, we propose an object-guided instance segmentation method to improve both detection and segmentation performance. The advantages of our methods are three-fold. First, we improve the object detection ability for box-based instance segmentation by adopting a center keypoint-based object detector. Second, we reuse the object features as guidance to make the model learn to separate target object from the neighboring objects within a cropped ROI region. Third, we improve the segmentation quality by densely sampling and refining the point-wise features around the boundary areas in an auxiliary manner. The proposed method is size-adaptive to all instances with different scales and sizes. We evaluate our method on three representative datasets, i.e., neural cell, plant phenotyping, and DSB2018. Experimental results demonstrate the superiority of our method compared to the state-of-the-art approaches. In addition, to further investigate the effectiveness of our method, we also apply it to the cell tracking and leaf segmentation challenges, and achieve decent performance.

\subsection{Conflict of Interest}
This paper is an extended version of our conference paper ``Object-Guided Instance Segmentation for Biological Images"  \cite{yi2019object}. In this manuscript, we additionally propose an auxiliary feature refinement module to further improve the segmentation performance. Besides, we add more equations, figures, comparison experiments and experimental results.

\subsection{Paper Organization}
The paper is organized as follows. In Section \ref{sec: methodology}, we present the proposed method in detail. In Section \ref{sec: experiments}, we quantitatively and qualitatively evaluate our method on three datasets. In Section \ref{sec: results and discussion}, we offer an in-depth analysis of the experimental results, and additionally apply our method to two instance segmentation challenges for further analysis. We conclude the paper in Section \ref{sec: conclusion}.

\section{Methodology}
\label{sec: methodology}
An overview of the proposed object-guided instance segmentation model is provided in Fig.~\ref{fig1}. Our end-to-end multi-task model contains two branches: (1) a center keypoint-based object detection branch (upper branch); (2) an object-guided image segmentation branch (lower branch). The auxiliary feature refinement module is built along with the segmentation branch. In the following sections, we first introduce our object detection branch in Section \ref{sec: object detection}. Then we describe the details of our instance segmentation strategy in Section \ref{sec: object segmentation}.

% input figure 2
\input{images/fig2}
%%%%%%%%%%%%%%%%%%%%%%%%%%%%%%%%%%%%

\subsection{Center Keypoint-Based Object Detection}
\label{sec: object detection}
We adopt a U-shaped architecture \cite{ronneberger2015u} for center keypoint-based object detection. The backbone network is built upon ResNet50 conv1-5 \cite{he2016deep} blocks, where skip connections are introduced to combine the high-level semantics in deep layers and low-level fine details in shallow layers. Batch Normalization is employed in the combination module B (see Fig.~\ref{fig1}). Finally, a detection head (DecHead in Fig.~\ref{fig1}) is built on the last feature map (c3 in Fig.~\ref{fig1}). We illustrate the details of detection head in Fig.~\ref{fig2}. As is shown, the detection head outputs three maps: heatmap, width-height map, and offset map. We use $L_{heat}$, $L_{wh}$, and $L_o$ to represent the loss of heatmap, width-height map, and offset map, respectively. The loss of detection branch is $L_{dec}=L_{heat}+L_{wh}+L_o$.

\subsubsection{Center Heatmap}
Center heatmap is a mask that contains 2D Gaussian blobs, and can be obtained as follows. First, we generate a binary mask where the positive locations indicate the cell centers and negative locations correspond to the background. However, directly using this binary mask as ground-truth for pixel classification is ineffective due to the severe imbalance between the number of positive and negative pixels. To address this problem, following the work of \cite{law2018cornernet,zhou2019objects}, instead of penalizing all the background pixels, we reduce the penalty for background points within a circular blob region. The motivation is that if the real center point is not captured, we still have enough candidate points within the circular blob. Moreover, the bounding box at the candidate point should have a high intersection over union (IOU) with the bounding box at the center point. Therefore, after obtaining the binary mask, we create the ground-truth center heatmap by placing an unnormalized 2D Gaussian blob, i.e., $e^{-\frac{(u-c_u)^2+(v-c_v)^2}{2\sigma^2}}$, around each positive point (see Fig.~\ref{fig2}), where $(c_u, c_v)$ is the object center (positive) location, $(u, v)$ is the background point in the blob, and $\sigma$ is $1/3$ of the radius. The radius is determined by ensuring the bounding box at the background point within the blob has a large overlap (e.g., $0.7$ IOU) with the ground-truth bounding box at the center point \cite{law2018cornernet}. We use a variant focal loss to optimize the parameters \cite{law2018cornernet,zhou2019objects}:
\begin{equation}
    L_{heat} =
    -\frac{1}{N}
    \begin{cases}(1-p_i)^\alpha \log(p_i)&\text{if }  y_i=1\\
    (1-y_i)^\beta(p_i)^\alpha\log(1-p_i)&\text{otherwise}
    \end{cases},
\end{equation}
where $i$ indexes the $i$-th location on the heatmap, $N$ is the total number of center points, $y$ is the ground-truth heatmap and $p$ refers to the predicted heatmap. $y_i=1$ or $y_i=0$ means the $i$-th location is a positive point (i.e., center point) or a negative point (i.e., background point), respectively. %In our experiments, 
When $y_i\in [0,1)$, we multiply the term ($1-y_i$) with the weight $(p_i)^\alpha\log(1-p_i)$ for a background point. As we can see, when a point is very close to the center point (i.e., $y_i$ is large), its penalty will be small (i.e., $1-y_i$ is small). On the contrary, if the a point is far away from the center point (i.e., $y_i$ is small), the penalty for the point is large (i.e., $1-y_i$ is large). Empirically we set $\alpha=2$, $\beta=4$  in this paper, following \cite{lin2017focal,law2018cornernet}. The predicted center heatmap is refined using a non-maximum-suppression (NMS) operation. This operation employs a $3\times 3$ max-pooling layer on the center heatmap. Finally, the center points are gathered according to their local maximum probabilities on the predicted heatmap. We refer to the heatmap probability at each center point as the confidence score of the corresponding decoded bounding box.

\subsubsection{Width-Height Map}  
After obtaining the object center locations, in the next step, we extract the box width and height parameters at the center points from the width-height map. The size of width-height map is the same as center heatmap. In width-height map,  only values at the center points participate in gradient computation and backpropagation. We use smooth$_{L_1}$ loss to regress the widths and heights of the bounding boxes at center points:
\begin{equation}
    L_{wh} = \sum_i\sum_{b\in\{w,h\}} \text{smooth}_{L_1}(b_i-\hat{b}_i),
\end{equation}
\begin{equation}
    \text{smooth}_{L_1}(\theta) = 
    \begin{cases}
    0.5\theta^2 & \text{if } |\theta|<1\\
    |\theta|-0.5 & \text{otherwise}
    \end{cases},
\end{equation}
where $i$ indexes the center points, $b$ and $\hat{b}$ refer to the predicted and ground-truth box parameters, respectively.

\subsubsection{Offset Map}
\label{sec: offset map}
As can be seen from Fig.~\ref{fig1}, the output map %size is not the same as 
is downscaled compared to the input image. For example, the map $c^3$ is downscaled compared to $c^1$. We could use a mirrored architecture like U-Net \cite{ronneberger2015u} to align the sizes of input and output maps. However, this turns out to be a sub-optimal strategy since the detection task does not require the fine details of objects, such as the contours. In particular, using a large output resolution for center detection would bring unnecessarily more computation and severe imbalance issues when training the heatmap. Therefore, in this paper we keep this non-mirrored architecture as in the general keypoint detection methods \cite{law2018cornernet,newell2016stacked}. When mapping the integer center points from input $c^1$ to the output $c^3$ (Fig.~\ref{fig1}), we will have a floating-point discrepancy between the downscaled value and the quantified value. We use an offset map to compensate this discrepancy at each center point. Suppose $n$ is the downsized factor, $(u,v)$ is a location in the input image. The offset map can be formulated as:
\begin{equation}
    o_i = (\frac{u_i}{n}-\lfloor\frac{u_i}{n}\rceil, \frac{v_i}{n}-\lfloor\frac{v_i}{n}\rceil),
\end{equation}
where $i$ indexes the $i$-th center point. In the training process, we apply smooth$_{L_1}$ loss to regress the offsets at center points:
\begin{equation}
    L_{o} = \sum_i \text{smooth}_{L_1}(o_i-\hat{o}_i),
\end{equation}
where $o$ and $\hat{o}$ refer to the predicted and ground-truth offsets at the center points, respectively.

\subsubsection{Bounding Boxes Decoding} With the predicted center heatmap, we extract center point locations $c=(c_{u}, c_{v})$ according to the local maximum probabilities on the heatmap. At the same time, we take out width $w$ and height $h$ at each center point from the width-height map, and offset $o=(o_{u}, o_{v})$ from the offset map. Then the compensated center location is $\Bar{c} = (c_{u}+o_{u}, c_{v}+o_{v})=(\Bar{c}_{u}, \Bar{c}_{v})$. Given the downsized factor $n$ (see Section \ref{sec: offset map}), the output bounding box is decoded as $(n\Bar{c}_{u}, n\Bar{c}_{v}, nw, nh)$.

\vspace{-0.5em}
\subsection{Object-Guided Segmentation}
\label{sec: object segmentation}
With the decoded bounding boxes from Section \ref{sec: object detection}, we crop the regions within bounding boxes out of feature maps $c^1$-$c^5$ (e.g., blue blocks in the upper branch of Fig.~\ref{fig1}) to perform a coarse-to-fine instance segmentation. We use grid sampling with bilinear interpolation to crop the ROI patches. We omit those ROIs when their widths or heights are smaller than 2 pixels. To perform segmentation, we build an object-guided segmentation branch (the lower branch in Fig.~\ref{fig1}) which employs skip connections (Module S in Fig.~\ref{fig1}) to combine the deep and shallow features. The shallow features from encoder layers $c^1$-$c^2$ contain rich morphology details such as leaf stalks. While being beneficial to recovering the fine details of objects, it also brings difficulty for the network to differentiate the target object from neighboring objects within the same cropped region.  To solve this problem, we propose to leverage the object information (layers $c^3$-$c^5$) from the detection branch and thereby build an object-guided segmentation branch.  Particularly, in this branch we adopt instance normalization to normalize the combined features. Given an ROI region $r\in \mathbb{R}^{H\times W}$, instance normalization is formulated as:
\begin{equation}
    r_{h,w}^\prime = \gamma(\frac{r_{h,w}-\mu}{\sigma})+\eta,
\end{equation}
where $\mu$ and $\sigma$ are the mean and variance of an ROI region, respectively, $\gamma$ and $\eta$ are two learned scaling factors for the network to control the extent of normalization. Note that we do not batch the cropped ROI patches, because the ROI patches are of different aspect ratios and scales, and simply padding and compiling them into a batch would introduce more computational cost. Therefore, we employ instance normalization instead of batch normalization in the combination module S (see Fig.~\ref{fig1}).

The segmentation head comprises two convolutional layers with kernel size $3\times3$. We denote the predicted segmentation mask by $x\in \mathbb{R}^{H\times W}$. The parameters are optimized using binary cross-entropy loss, which is formulated as follows:
\begin{equation}
    L_{seg} = -\sum_i (\hat{x}_i\log x_i + (1 -\hat{x}_i)\log (1-x_i)),
\end{equation}
where $i$ indexes the pixels on map $x$, and $\hat{x}$ represents the ground-truth mask. 

To ensure the cropped ROI patches are not misaligned, we perform Hough Voting  \cite{papandreou2018personlab} to assign the floating-point value to its four surrounding integer grid points. Next, we conduct binary thresholding to the mask. Finally, we resize the mask to the original image size through nearest-neighbor interpolation.

\vspace{-0.5em}
\subsection{Auxiliary Feature Refinement Module}
In this section, we introduce the auxiliary feature refinement module, which is an extension of our previous work \cite{yi2019object}.

Image segmentation can be viewed as a pixel labeling task performed on regular grids. In contrast to the pixels in smooth regions (i.e., regions inside the object boundaries), pixels on the object boundaries are usually insufficiently learned due to the regular sampling schemes. One existing work \cite{ronneberger2015u} deals with this problem by assigning more weights to the boundary pixels. However, its learning process is still based on the regular grid of pixels. In this paper, we propose an auxiliary feature refinement module, which densely samples and refines the boundary region features in the training process.  The refined features (i.e., features in $e^1$ and $e^2$) are shared with the segmentation head (see Fig.~\ref{fig1}). The sampling points of our method are non-uniformly distributed and are floating-point values. In this manner we provide more accurate information (e.g., labels are interpolated at floating-point positions) in the training process and therefore enable better learning of the boundary. In the following sections, we first introduce the generation of non-uniform sampling points, and then we illustrate the auxiliary feature refinement module.

%%%%%%%%%%%%%%%%%%
\input{images/fig3}
%%%%%%%%%%%%%%%%%%%%
 
\subsubsection{Non-Uniform Sampling Points} 
Non-uniform sampling points are employed to extract the point-wise features in the boundary regions. To create non-uniform sampling points, we first need to determine the boundary regions from the predicted mask. An example of the predicted mask is shown in Fig.~\ref{fig3}. The yellow and blue colors respectively represent the foreground and background regions where the network is confident. In other words, the network outputs very high probabilities for pixels in yellow regions and very low probabilities for pixels in blue regions. The green color reflects regions where the network is uncertain (i.e., the probability is around 0.5). We observe that these uncertain regions distribute in the high-frequency area, such as the boundary of objects. Based on this observation, we create an uncertainty map to identify the uncertain regions (e.g., boundary). Given a predicted mask $x\in \mathbb{R}^{H\times W}$, the uncertainty map is calculated by \cite{kirillov2019pointrend}:
 \begin{equation}\label{eq:uncertain map}
     x^\prime = -|2x-1|,
 \end{equation}
where the negative sign is introduced for the sorting purpose so that the values at uncertain regions correspond to the max.  With uncertainty map $x^\prime$ we can generate the non-uniform sampling points as follows. First, we randomly generate $kN$ (i.e., $k\times N$) sampling points whose coordinates are floating-point numbers, where $N=H\times W/D$. We use $D=8$ in this paper considering the computation cost. To be specific, these points are randomly generated on the mask, where the horizontal and vertical coordinates are random floating-point numbers. Next, we sort these points according to their values on map $x^\prime$ in descending order. The values are obtained through bilinear interpolation. Finally, we select the top $\beta N$ uncertain points as the sampling points.  We use $k=3$ and $\beta=0.75$ as in PointRend \cite{kirillov2019pointrend}. In Fig.~\ref{fig3} we visualize the non-uniform sampling points, which are shown as red points.
 
\subsubsection{Auxiliary Feature Refinement}
After obtaining the non-uniform sampling points, we extract the corresponding point-wise features (e.g., a 64-dimension vector per pixel) from $e^1$ and $e^2$ (see Fig.~\ref{fig3}) through bilinear interpolation. The extracted point-wise features are then fed into the refinement head, which combines and refines these features using two $1\times 1$  convolutional layers. We use binary cross-entropy as the training loss for feature refinement head:
\begin{equation}
    L_{ref} = -\sum_i (\hat{\zeta}_i\log \zeta_i + (1 -\hat{\zeta}_i)\log (1-\zeta_i)),
\end{equation}
where $i$ indexes the pixels on feature maps $e^1$ and $e^2$, $\zeta$ and $\hat{\zeta}$ refer to the predicted logits at the sampling points and the ground-truth point labels, respectively.  The ground-truth point labels $\hat{\zeta}$ are soft labels which are obtained through bilinear interpolation at the floating-point sampling positions on the ground-truth mask $\hat{x}$.

\section{Experiments}
\label{sec: experiments}
\subsection{Datasets}
We evaluate the performance of our method on three biological datasets: neural cell, plant phenotyping, and DSB2018. The first two datasets contain instances with long and slender structures, while the last dataset consists of instances that generally have regular round shapes.

\subsubsection{Plant Phenotyping}
The study of plant phenotyping under different environmental conditions is critical to the understanding of plant function \cite{minervini2016finely}. The phenotype relies on plant's fine-grained properties, such as leaf numbers, shapes, age, maturity level, and so on. By capturing individual leaf objects, instance segmentation serves as an important tool to analyze these properties quantitatively. In this paper, we apply the proposed method to a public plant phenotyping dataset \cite{minervini2016finely}. This dataset contains 473 top-down view plant images with various image sizes. We split the images into three subsets: 284 images for training, 95 images for validation, and 94 images for testing. The leaves in this dataset generally have long and slender stalks. Besides, the boundaries between leaves are blurry, making instance segmentation even more challenging.

\subsubsection{Neural Cell} 
The cellular mechanisms about the specification of neurons, astrocytes, and oligodendrocytes from a single stem cell remain mysteries in neural science \cite{ravin2008potency}. The study of interactions among neural cells plays an important role in understanding such cellular mechanisms. However, it is non-trivial to capture the filopodia and lamellipodia of neural cells, where the interactions typically happen. These structures are extremely tiny and are in low-contrast compared against the background in a given image. In this paper, we apply our method to a neural cell dataset, where the images are sampled from a collection of time-lapse microscopic videos of rat CNS stem cells \cite{ravin2008potency}. This neural cell dataset contains 644 gray-scale images with size 512$\times$640. We randomly select 386 images for training, 129 images for validation, and 129 images for testing. The neural cells have extremely irregular shapes with long and slender protrusions, and thereby bring difficulties for instance segmentation.

\subsubsection{DSB2018} 
Instance segmentation of cells' nuclei offers an important tool for researchers to measure how cells respond to various drug treatments and understand the underlying biological processes. In this paper, we apply our method to a public cell nuclei dataset, DSB2018, which is obtained from the training set of the 2018 Data Science Bowl. The task of this dataset is to capture the nuclei in diverse images. The images in the dataset are acquired under different conditions. The cell nuclei are mostly in round shapes, and their sizes vary among different cell types. We randomly split the total 670 images into three subsets: training (402 images), validation (134 images), and testing (134 images). 

\subsection{Evaluation Metric}
We use AP (average precision) \cite{everingham2011pascal} as a metric to evaluate both object detection and instance segmentation performances, following the works of \cite{he2017mask,chen2019tensormask,yi2019object}. AP metric summarizes the precision-recall curve at a certain IOU threshold $\alpha$. We use bounding box IOU between the predicted bounding box and the ground-truth bounding box when evaluating the detection performance (AP$^{dec}$). Similarly, we use mask IOU between the predicted mask and the ground-truth mask to evaluate the instance segmentation performance (AP$^{seg}$). 

To compute AP, first, we sort the predicted bounding boxes or masks according to their confidence scores over all testing images. As mentioned before, the confidence score for our method refers to the probability that an extracted point is the object center, and can be obtained from center heatmap. In the next step, we calculate the IOU between predicted bounding box or mask $p$ and its ground-truth $\hat{p}$:
\begin{equation}
\label{Eq:IOU}
    \text{IOU} = \frac{p\cap \hat{p}}{p\cup \hat{p}}.
\end{equation}
We consider a predicted $p$ as true positive (TP) if its IOU with $\hat{p}$ is greater than a given threshold $\alpha$. For a ground-truth $\hat{p}$, it is matched to a predicted $p$ that has the highest IOU with $\hat{p}$ among all the true positive predictions; the remaining true positive predictions are considered as false positive (FP). If a ground-truth $\hat{p}$ has no matched prediction $p$ satisfying IOU$\ge\alpha$, we record it as false negative (FN). After obtaining the TP, FP, and FN, we calculate the precision-recall curve. The AP$@\alpha$ is calculated by measuring the area under the precision-recall curve.  In this paper, we measure the  AP at 10 IOU thresholds ranging from 0.5 to 0.95 with an interval of 0.05. We report the average value over the 10 thresholds as AP$_{\{0.5:0.05:0.95\}}$.

\input{images/fig4}
\input{tables/table0}
%\input{tables/table1}

\input{images/fig5}

\subsection{Implementation Details}
We implement our model with PyTorch1.2 on NVIDIA Quadro RTX 6000 GPUs. For all datasets, the input images are resized to $512\times512$ before being fed into the network. The number of instances on each image is not fixed. We augment the training images with random flipping and cropping. The parameters of the backbone network are pretrained on ImageNet. Other weights of the network are initialized using standard Gaussian distribution. We train the network using Adam optimizer \cite{kingma2014adam} with an initial learning rate $1.25\times10^{-4}$. We use Detectron2 to implement PointRend \cite{kirillov2019pointrend}.

\section{Results and Discussion}
\label{sec: results and discussion}

In this section, we compare the proposed method with two box-free instance segmentation methods (i.e., DCAN \cite{chen2017dcan} and Cosine Embedding \cite{payer2018instance}) and three box-based instance segmentation methods (i.e., Mask R-CNN \cite{he2017mask}, PointRend \cite{kirillov2019pointrend}, and Keypoint Graph \cite{YiMICCAI2019}). 
The quantitative and qualitative results are shown in Table~\ref{table1} and Fig.~\ref{fig4}. In Table~\ref{table1}, $\text{AP}^{dec}$ refers to averaged detection performance over the 10 bounding box IOU thresholds. $\text{AP}^{seg}$ is the averaged instance segmentation performance over the 10 mask IOU thresholds. We provide AP$^{seg}$ at mask IOU thresholds of 0.5 and 0.75 in Table~\ref{table1}.  

\subsection{Box-Free Instance Segmentation}
DCAN and Cosine Embedding are two box-free methods, which infer the object masks without bounding box detection. The box-free methods are prone to errors due to blurred boundaries or uneven textures. For example, as shown in Table~\ref{table1}, the detection and instance segmentation performances of DCAN are poor. The reason can be found in Fig.~\ref{fig4}c, where DCAN fails to separate the overlapping leaves due to their unclear boundaries. Besides, by fusing the contour information with semantic maps, the predicted masks would lose details such as leaf stalks and cell protrusions. Cosine Embedding performs even worse compared to DCAN. In Table~\ref{table1}, Cosine Embedding is about 10 points lower than DCAN in AP$^{seg}$. From Fig.~\ref{fig4}d, we observe that Cosine Embedding tends to produce fragmentary masks within the same object, resulting in more false positive cases. This phenomenon reveals that segmentation by clustering would be severely affected by the uneven textures of objects. 

\subsection{Box-Based Instance Segmentation}
\label{sec: Box-Based Instance Segmentation}
Compared with box-free instance segmentation methods, box-based strategies excel in identifying individual objects. For example, as shown in Table~\ref{table1}, box-based methods (i.e., Mask R-CNN, PointRend, Keypoint Graph, and Ours) have better AP$^{dec}$ than box-free methods (i.e., DCAN and Cosine Embedding). Among these box-based methods, the keypoint-based detection strategy (i.e., Keypoint Graph and Ours) overall has better detection accuracy AP$^{dec}$ than the anchor box-based strategy (i.e., Mask R-CNN and PointRend), indicating that keypoint-based detectors are more preferable for instance segmentation due to their improved detection performance. On the other hand, the improvement of segmentation quality in turn boosts the detection performance. For example, compared with Mask R-CNN, PointRend achieves better detection accuracy especially on plant and neural cell datasets even though PointRend and Mask R-CNN adopt the same object detector. The reason is that PointRend renders the boundary pixels for the output mask and thereby preserves more boundary details; in this way, it further refines the bounding box locations. From Fig.~\ref{fig4}e we observe that Mask R-CNN can hardly capture the object details such as cell protrusions and leaf stalks. The reason would be that Mask R-CNN uses ROI-Align to pool the ROI regions and fix the ROI resolution to $28\times 28$. Such a fixed resolution would not be enough to capture the object details. On the contrary, PointRend (Fig.~\ref{fig4}f) computes the output mask iteratively up to a high resolution ($224\times 224$) and it captures more boundary details than Mask R-CNN. As shown in Table~\ref{table1}, PointRend achieves about 10 points higher AP$^{seg}$ than Mask R-CNN on the plant and neural cell datasets, where the objects contain more fine details. However, from Fig.~\ref{fig4}f we observe that PointRend would generate non-smooth (i.e., broken) segmentation masks especially for the long and slender protrusions of neural cells. The reason would be that PointRend only renders points in the boundary area and combines the rendered result with the initial coarse segmentation mask. This process does not explicitly capture the relationship between boundary and interior areas, leading to incomplete segmentation results (which may contain holes and fragments). Different from PointRend, our method  refines the boundary features in an auxiliary way and therefore produces a more complete output mask (see Fig.~\ref{fig4}i). 

\subsection{Effectiveness of Object-Guided Segmentation}
Different from Mask R-CNN and PointRend which use anchor box-based detector, Keypoint Graph predicts bounding boxes by detecting and grouping their corner and center points. Without the imbalance issue in the anchor box-based methods, Keypoint Graph tends to achieve better detection performance. Also, it can better capture the object details due to the coarse-to-fine segmentation branch. In Table~\ref{table1}, we observe that the performance gap between Mask R-CNN and Keypoint Graph is small ($<$10 points in AP$^{seg}$) on DSB2018 and plant datasets. In contrast, on neural cell dataset which contains more long and slender fine structures, the gap is large ($>$20 points in AP$^{seg}$). However, Keypoint Graph can hardly identify the small objects (see Fig.~\ref{fig3}g) due to the overlapping of box keypoints. Besides, it is unable to suppress the neighboring objects effectively (see the failure cases indicated by the white arrows in Fig.~\ref{fig4}g). This kind of segmentation failure is even more evident on the plant dataset. In particular, from Table~\ref{table1} we can observe that Keypoint Graph performs about 10 points lower in AP$^{seg}$  than PointRend on the plant dataset. The weakness that the model cannot distinguish the target from its neighboring objects within the same cropped ROI region is difficult to overcome for the coarse-to-fine segmentation strategy. The reason is that coarse-to-fine segmentation consumes low-level features from the shallow layers; these features typically contain rich texture and boundary information and would bring difficulty for the network to differentiate the target. Note that Mask R-CNN and PointRend do not have this problem because they do not use shallow feature layers. However, Mask R-CNN tends to lose the fine details of objects and PointRend attempts to address this issue by rendering the boundary pixels.

Different from previous works, we solve this problem in a simple yet effective manner. In particular, we use the object features as guidance to make the model learn to suppress the neighboring objects within the same ROI region. As shown in Fig.~\ref{fig4}h, the proposed method is able to separate the overlapped leaves in the plant dataset, and suppress the neighboring cells in the neural cell dataset compared to Keypoint Graph. In addition, the proposed center keypoint-based detection strategy enables the network to better capture the small objects, as shown in the second column of Fig.~\ref{fig4}h.

\input{images/fig6}

\subsection{Effectiveness of Auxiliary Feature Refinement Module}
To demonstrate the effectiveness of auxiliary feature refinement module, we conduct ablation studies and consider two versions of our method: Ours W.O. (without) Refine and Ours W. (with) Refine. From Table~\ref{table1} we observe that Ours W.O. Refine alone achieves the best performance in both detection and segmentation, compared to the baseline methods. With auxiliary feature refinement module, Ours W. Refine boosts the segmentation performance further, especially on the plant and neural cell datasets.  From Fig.~\ref{fig4}h-i, we observe that Ours W. Refine better captures the boundary details of neural cells (indicated by thick arrows), compared with Ours W.O. Refine. We additionally illustrate the AP$^{seg}$ at different IOU thresholds in Fig.~\ref{fig5}. It can be observed that the proposed method achieves the best performance compared to the state-of-the-art methods. Also, Ours W. Refine performs better than Ours W.O. Refine especially for the plant and neural cell datasets. The reason would be that the objects in these two datasets contain more high-frequency and low-contrast areas, such as the boundaries of protrusions. The auxiliary feature refinement module learns to pay more attention to these areas and therefore boosts the overall performance.

We also conduct ablation studies with regard to different sampling strategies for the auxiliary feature refinement module (see Table \ref{table2}). We define eight sampling strategies. Given $N$ we randomly generate $kN$ sampling points on the uncertainty map (Eq.~\ref{eq:uncertain map}). Then the uniform strategies mean that a total of $\beta N=kN$ points are utilized for feature refinement. For biased strategies, we select the top $\beta N$ uncertain points from the whole $k N$ points for feature refinement (see Fig.~\ref{fig6}). As can be seen from Table \ref{table2}, for DSB2018 dataset, the uniform and biased strategies almost have no impact on the performance. The reason would be that the cell nuclei are generally in round shapes and contain less fine structures such as cell protrusions and leaf stalks, and therefore are less challenging to segment. As a result, the proposed auxiliary feature refinement module does not show obvious advantages on DSB2018 dataset. For the plant phenotyping dataset that has overlapping leaves and long-tail structures, the biased strategies perform better than the uniform ones, indicating that focusing on the uncertain regions (e.g., boundary) helps network improve the segmentation quality. Further increasing the level of bias (i.e., increasing $k$) does not make obvious improvement. For neural cell dataset, the moderately biased strategy performs the best. Notice that the uniform sampling strategy ($k=1, \beta=1$ or $k=3, \beta=3$) performs $\sim 1$ point lower than the moderately biased sampling strategy ($k=3, \beta=0.75$) on neural cell and plant datasets. This suggests the biased sampling is better than uniform sampling. Thus, we finally choose $k=3$ and $\beta=0.75$ as they achieve the best performance on neural cell and plant datasets.

\input{tables/table2}

\subsection{Inference Speed}
In Table~\ref{table1} we compare the efficiency of different methods and report their inference speeds (FPS, frame per second). As shown in Table~\ref{table1}, PointRend achieves the fastest speed due to its efficient implementation based on Detectron2 as well as the boundary-specific rendering strategy. However, as shown in Fig.~\ref{fig4}, PointRend would generate incomplete instance masks as it only renders the boundary pixels. Compared to PointRend, the proposed method is less efficient but it is still faster than other competitive methods. The reason would be that the proposed method adopts a single-stage object detector, which is more efficient than the two-stage object detector used in Mask R-CNN. Besides, our center keypoint-based object detector only detects one keypoint for each object and thereby avoids the complicated post-grouping process required by Keypoint Graph. Finally, the non-mirrored detection architecture (Fig.~\ref{fig1}) makes the inference even faster.

\subsection{Evaluation on Cell Tracking Challenge}
\label{sec:celltrack}
To further investigate the effectiveness of our method, we evaluate our method on the external Cell Segmentation Benchmark of Cell Tracking Challenge (CTC) \cite{ulman2017objective}. In particular, we evaluate our method on three datasets: Fluo-N2DL-HeLa, PhC-C2DH-U373, and PhC-C2DL-PSC. We use the same training configurations as before and only use random crop and flipping for data augmentation. The model is trained separately for different datasets. For datasets Fluo-N2DL-HeLa and PhC-C2DH-U373, we only use the silver segmentation annotations (which are computer-generated) for training as the cells are in regular shapes and the gold annotations (which are human-annotated) are incomplete and limited in number. For PhC-C2DL-PSC which has long-tail structures, we finetune our model with gold annotations for 10 epochs.  The performance is evaluated via online server. Three metrics are used: (1) DET, which is defined as normalized Acyclic Oriented Graph Matching (AOGM-D) measure\cite{matula2015cell} for detection; (2) SEG, which is the IOU between predicted and ground-truth masks; (3) OP$_{\text{CSB}} = 0.5$(DET + SEG). We report the evaluation results in Table \ref{table3}. We observe that although there are some gaps between our method and the top winners' methods, our method still achieves decent performance. We would like to note that, winning a challenge depends on many factors, such as dataset-specific designs. Besides, the lack of training data ($\sim$200 images for each dataset) limits the power of deep neural networks which contain millions of parameters (62M for our method). Moreover, the annotations of the three datasets are incomplete. For example, the cells in PhC-C2DL-PSC contain many long-tail structures but the silver annotations fail to capture them (see Fig.~\ref{fig7}). These incomplete silver annotations would mislead the model and make it ignore these fine boundary structures. In contrast, the gold annotations capture the fine details more accurately. However, the number of these annotations is very limited (e.g., only 4 images on the PhC-C2DL-PSC dataset). This means that the gold annotations are far from enough to teach the model to identify the cell fine structures correctly. Finally, the data in these datasets are actually videos, and our method is not designed to work on such data. In the future we plan to exploit the temporal information of videos, which we believe would further benefit the performance of our method.

\input{images/fig7}

\input{tables/table3}
\input{tables/table4}

\subsection{Evaluation on Leaf Segmentation Challenge}
\label{sec:LSC-challenge}
Finally, we also evaluate our method on the Leaf Segmentation Challenge (LSC) \cite{minervini2016finely,scharr2014annotated}. In this challenge, there are five datasets A1-A5, where A5 contains images from A1-A4. From Fig.~\ref{fig8} we can see A2 contains long and slender leaf stalks, A3 contains larger leaves, and A4 contains smaller leaves. All the datasets, including the training and testing, have both raw images and foreground binary annotations. For consistent and fair evaluation, we do not use the foreground binary annotations. During training we use random crop and flipping for data augmentation. We only train one model using all the datasets A1-A4. The metric DiffFG is the difference in object count. absDiffFG is the absolute value of DiffFG. FgBgDice is the DICE on the foreground binary mask. From Table \ref{table4} we observe that our method performs the best on A1 dataset and the worst on A3 dataset. The reason would be that compared to other datasets, A3 contains larger leaves and different types of plants (see Fig.~\ref{fig8}), and its number of training images is small (27 images). For the hybrid dataset A5, our method achieves decent performance.

\subsection{Discussion of Adverse Effect}
Although the proposed method competes favorably with the state-of-the-art works, there still exist several adverse effects.
\subsubsection{Fragmentary Segmentations}
As shown in Fig.~\ref{fig4}, for neural cells, our approach sometimes would generate fragmentary segmentations. The low-contrast image regions and abnormal shapes of the neural cells would be the reasons. To address this issue, we plan to exploit shape priors in our future work.

\input{images/fig8}
\subsubsection{Annotation Hungry}
As discussed in Subsection \ref{sec:celltrack} and \ref{sec:LSC-challenge}, our approach is not very effective when the number of well-annotated images is small. For example, our model would fail to capture the large-scale objects when the number of corresponding training images is very limited. To address this weakness, we would like to explore the few-shot learning and unsupervised learning methods in the future.

\section{Conclusion}
\label{sec: conclusion}
In this paper, we propose a novel object-guided instance segmentation method. The proposed method reuses the object features as guidance to make the model focus on the target and suppress its neighboring objects within the same cropped ROI region. The proposed auxiliary feature refinement module enables the network to refine the boundary features and further boosts the segmentation quality. Over different datasets we show that our method achieves competitive results compared to the state-of-the-art methods.

% \bibliographystyle{IEEEtran}
% \bibliography{tmi}
\input{main.bbl}

\end{document}

%% file: images/fig1.tex
% Figure 1%%%%%%%%%%%%%%%%%%%%%%%%%%%%%%%%%%%%%%%%%%%%%%%%%%%
% Architecture
\begin{figure}[!htb]
\includegraphics[width=0.98\columnwidth]{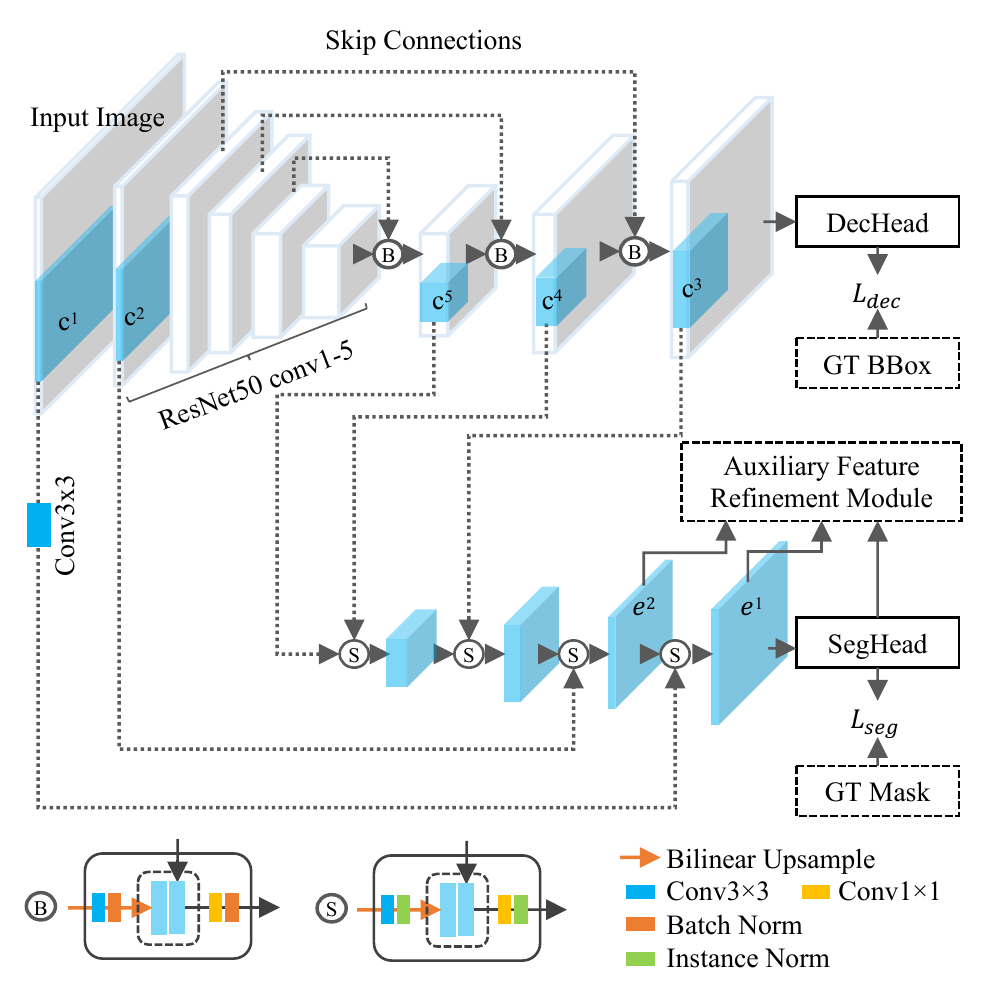}
\caption{The framework of our object-guided instance segmentation method. The backbone network comprises ResNet50 conv1-5 \cite{he2016deep}. We build two branches for object detection (upper branch) and segmentation (lower branch). Between the two branches, we employ skip connections (i.e., modules B and S) to combine the upsampled deep feature maps and shallow feature maps to enable the sharing of semantics. DecHead and SegHead refer to the detection head and segmentation head, respectively. GT is the abbreviation for ground-truth. The ROI patches (i.e., blue blocks) are cropped from $c^1$-$c^5$ to perform a coarse-to-fine instance segmentation. We use grid sampling with bilinear interpolation to crop ROI patches. The object features $c^3$-$c^5$ are utilized in the segmentation branch as a guide to help network suppress the undesired neighboring objects within the same ROI patch. The auxiliary feature refinement module samples the non-uniform floating-number points from $e^1$ and $e^2$ in the training process to refine the features in uncertain areas. As the cropped ROI patches have variant scales and aspect ratios, we use instance normalization in the segmentation branch.
} 
\label{fig1}
\vspace{-1em}
\end{figure}

%% file: images/fig2.tex
% Figure 2%%%%%%%%%%%%%%%%%%%%%%%%%%%%%%%%%%%%%%%%%%%%%%%%%%%
% Detection
\begin{figure}[!t]
\includegraphics[width=0.98\columnwidth]{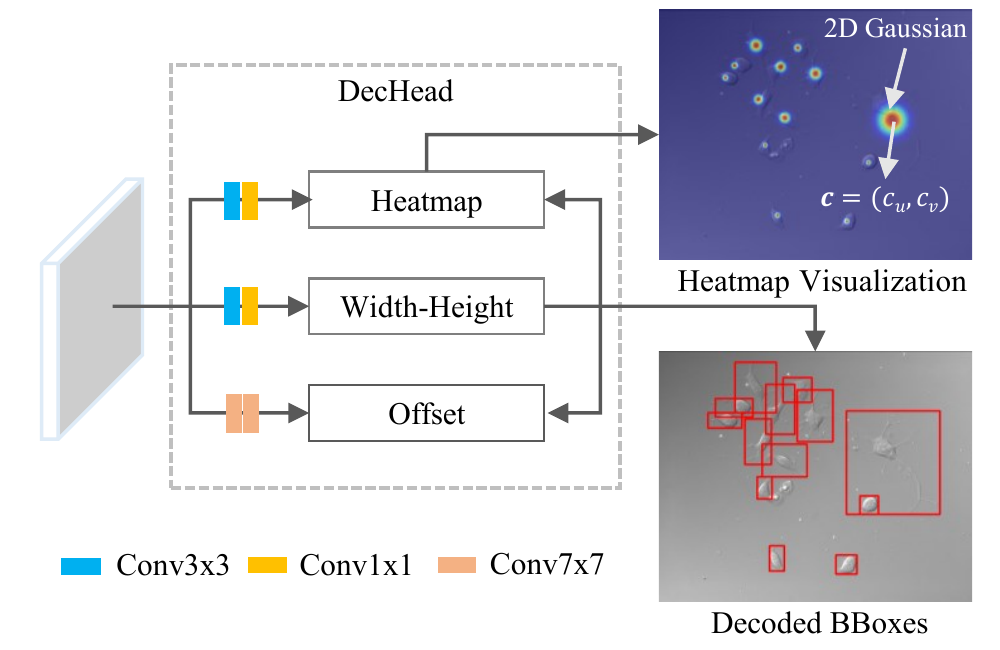}
\caption{The details of center keypoint-based object detection head. The detection head outputs three maps: heatmap, width-height map, and offset map. The symbol $c$ represents the bounding box center point of an object. We place a 2D Gaussian around each object center to generate the ground-truth heatmap. The heatmap, width-height map, and offset map are combined together to decode the object bounding boxes.} 
\label{fig2}
\vspace{-1em}
\end{figure}
%%%%%%%%%%%%%%%%%%%%%%%%%%%%%%%%%%%%%%%%%%%%%%%%%%%%%%%%%%%%

%% file: images/fig3.tex
% Figure 3%%%%%%%%%%%%%%%%%%%%%%%%%%%%%%%%%%%%%%%%%%%%%%%%%%%
% segmentation
\begin{figure}[!tbh]
\includegraphics[width=0.95\columnwidth]{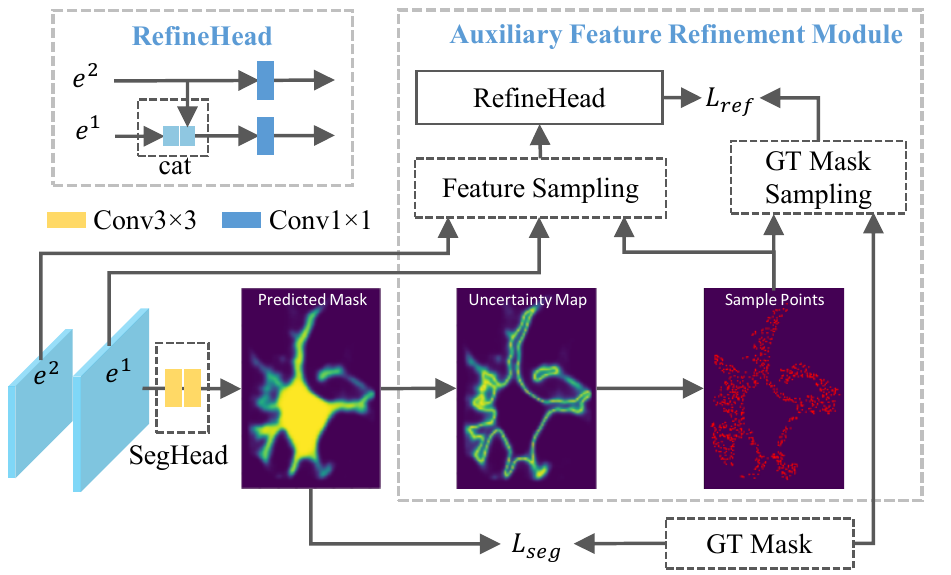}
\caption{Illustration of segmentation head and auxiliary feature refinement module. Feature maps $e^1$ and $e^2$ correspond to the last two layers of the segmentation branch (see Fig.~\ref{fig1}). The predicted mask is generated from the segmentation branch, where the yellow and dark blue colors represent the confident foreground and background regions, respectively. The green color reflects regions where the network is uncertain (i.e., the probability is around 0.5). The uncertain regions mainly distribute in the high-frequency areas (e.g., the boundaries). In the auxiliary feature refinement module, the feature sampling module generates the non-uniform sampling points in the uncertain regions and then densely samples the point-wise features (e.g., a 64-dimension  vector per pixel) from $e^1$ and $e^2$. Finally, RefineHead refines these point-wise features. The non-uniform sampling points provide more accurate information in the training process and therefore enable better learning of the boundary.} 
\label{fig3}
\vspace{-1em}
\end{figure}
%%%%%%%%%%%%%%%%%%%%%%%%%%%%%%%%%%%%%%%%%%%%%%%%%%%%%%%%%%%%

%% file: images/fig4.tex
% Figure 4%%%%%%%%%%%%%%%%%%%%%%%%%%%%%%%%%%%%%%%%%%%%%%%%%%%
%
\begin{figure*}[!t]
\centering
\includegraphics[width=0.95\textwidth]{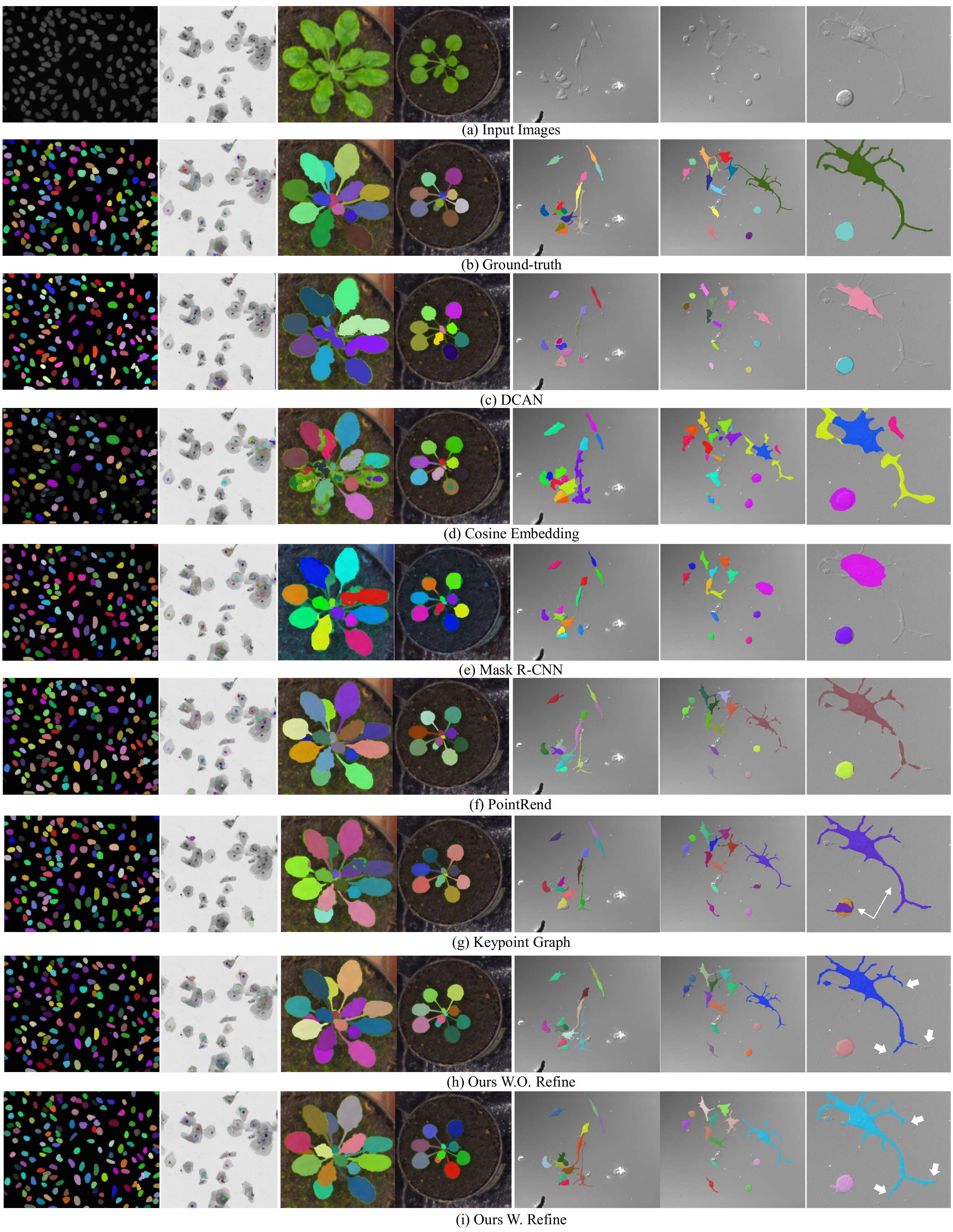}
\caption{Visualization of instance segmentation results. The output segmentation masks are projected onto the input images. The last column shows the cropped and enlarged images from the second last column. The thin white arrows point to a failure case where the Keypoint Graph  \cite{YiMICCAI2019} cannot differentiate the two objects within the same ROI region. The thick white arrows demonstrate the effectiveness of our auxiliary feature refinement module; in particular, the details of the boundary are preserved.} 
\label{fig4}
\end{figure*}
%%%%%%%%%%%%%%%%%%%%%%%%%%%%%%%%%%%%%%%%%%%%%%%%%%%%%%%%%%%%

%% file: tables/table0.tex
\begin{table*}[tbh!]
\centering
\caption{Quantitative instance segmentation results. Speed (FPS: frame per second) is measured on a single NVIDIA GeForce GTX 1080 GPU. AP is measured using Pascal VOC2010 metric \cite{everingham2011pascal}. The symbol ``--" denotes that the inference time is greater than 1min per image.}
\begin{tabular}{l|c|c|c|c|c|c|c}
\hline
% DSB Dataset %%%%%%%%%%%%%%%%%%%%%%%%%%%%%%%%%
Method &Datasets & Input Size & AP$^{dec}_{\{0.5:0.05:0.95\}}$ & AP$^{seg}_{\{0.5:0.05:0.95\}}$ & AP$^{seg}_{0.5}$ & AP$_{0.75}^{seg}$ & FPS
 \\ \hline
DCAN \cite{chen2017dcan} & \multirow{5}{*}{DSB2018} &$512\times 512$ &16.47&19.70&48.36&15.56  &2.67 \\
Cosine Embedding \cite{payer2018instance}& &$512\times 512$&2.25&3.36&14.96&0.24&  $-$\\
Mask R-CNN \cite{he2017mask} & &$512\times 512$&42.13&42.58&73.94&45.05 &1.01\\
PointRend \cite{kirillov2019pointrend}  & &$512\times 512$&40.10&41.08&75.02&41.56& 14.67\\
Keypoint Graph  \cite{YiMICCAI2019} & &$512\times 512$&49.07&50.63&76.13&56.72 & 1.54\\
Ours W.O. Refine&&$512\times 512$&50.41&61.14&84.85&65.14 &3.22\\
Ours W. Refine &  &$512\times 512$&\textbf{50.95}& \textbf{61.25}&\textbf{85.56}& \textbf{65.22} &2.80 \\
%%% Plant Dataset %%%%%%%%%%%%%%%%%%%%%%%%%%%%%%%%%%%%%%%%%%
\hline
DCAN \cite{chen2017dcan} & \multirow{5}{*}{Plant}&$512\times 512$ &7.03&16.67&38.86&13.02&12.99\\
Cosine Embedding \cite{payer2018instance}& &$512\times 512$&5.04&6.68&20.20&3.24& $-$\\
Mask R-CNN \cite{he2017mask} &&$512\times 512$&47.44&46.57&81.56&49.55 &5.57\\
PointRend \cite{kirillov2019pointrend} &&$512\times 512$&55.31&58.54&87.89&65.11& 16.87\\
Keypoint Graph  \cite{YiMICCAI2019}& &$512\times 512$&50.93&49.70&82.71&51.27& 1.82\\
Ours W.O. Refine&&$512\times 512$&59.45&74.11&92.20&79.15&5.45\\
Ours W. Refine & &$512\times 512$ & \textbf{62.99}& \textbf{76.71}&\textbf{92.98}&\textbf{81.08} &  5.69\\
%%%% Neural Dataset %%%%%%%%%%%%%%%%%%%%%%%%%%%%%%%%%%%%%%%%
\hline
DCAN \cite{chen2017dcan} & \multirow{5}{*}{Neural Cell}&$512\times 512$ &1.70&10.98&44.82&1.00& 4.87\\
Cosine Embedding \cite{payer2018instance}& &$512\times 512$&4.29&1.98&11.71&0.9& $-$\\
Mask R-CNN \cite{he2017mask} & &$512\times 512$&19.73&21.43&57.65&9.84& 1.13\\
PointRend \cite{kirillov2019pointrend} &&$512\times 512$&39.93&39.60&87.19&25.65& 15.97\\
Keypoint Graph \cite{YiMICCAI2019} &&$512\times 512$ &43.68&42.77&85.11&35.94& 1.86\\
Ours W.O. Refine&&$512\times 512$&\textbf{44.34}&58.26&94.33&67.24&5.11\\
Ours W. Refine &  &$512\times 512$  & 43.85 &\textbf{60.38}&\textbf{96.40} &\textbf{72.86} &5.70\\
\hline
\end{tabular}
\label{table1}
\vspace{-1em}
\end{table*}

%% file: images/fig5.tex
% Figure 5%%%%%%%%%%%%%%%%%%%%%%%%%%%%%%%%%%%%%%%%%%%%%%%%%%%
% 
\begin{figure*}[!t]
\centering
\includegraphics[width=0.95\linewidth]{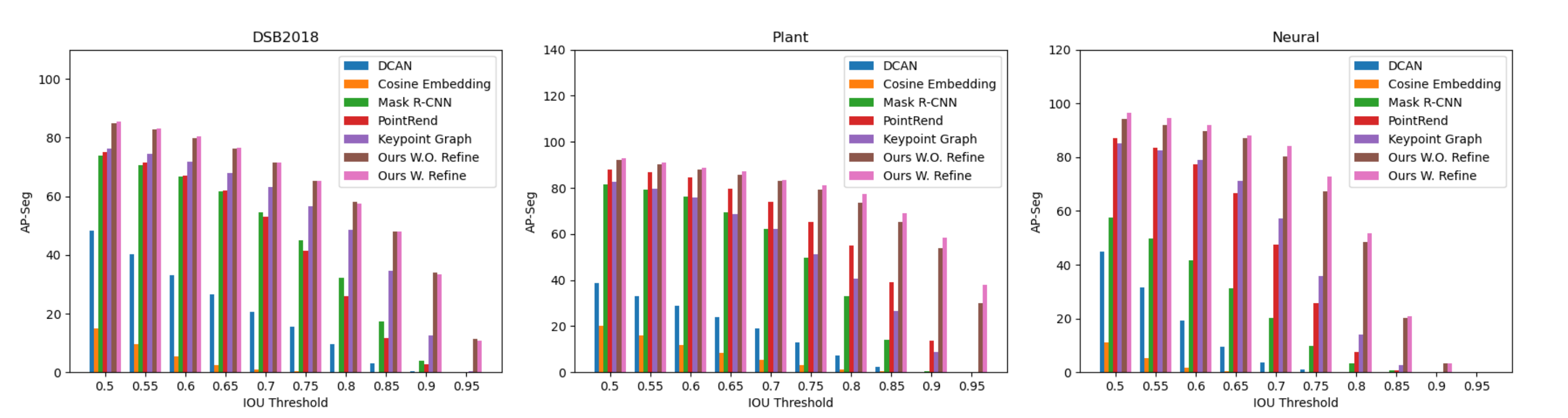}
\caption{Comparison of AP$^{seg}$ at different IOU thresholds (ranging from 0.5 to 0.95 with an interval of 0.05) for three datasets: DSB2018, plant and neural. Our method (i.e., Ours W. Refine) exhibits noticeable benefits on both plant and neural datasets, where the objects have protrusions or long-tail structures. } 
\label{fig5}
\vspace{-1em}
\end{figure*}
%%%%%%%%%%%%%%%%%%%%%%%%%%%%%%%%%%%%%%%%%%%%%%%%%%%%%%%%%%%%

%% file: images/fig6.tex
% Figure 6%%%%%%%%%%%%%%%%%%%%%%%%%%%%%%%%%%%%%%%%%%%%%%%%%%%
\begin{figure}[!t]
\centering
\includegraphics[width=0.9\columnwidth]{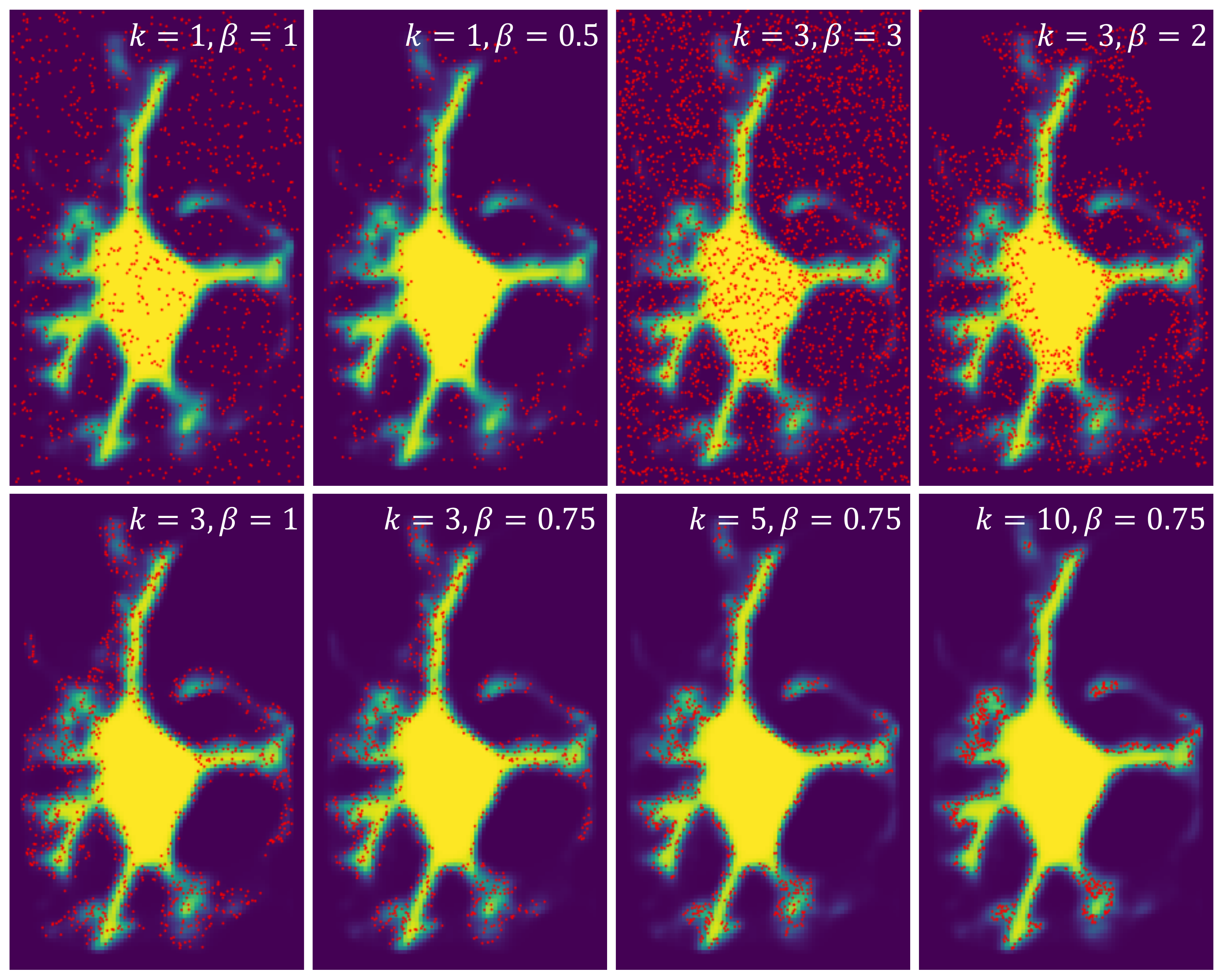}
\caption{Visualization of sampling points (red points) from different sampling strategies for the auxiliary feature refinement module. The points are scattered on a predicted mask.} 
\label{fig6}
\vspace{-1em}
\end{figure}
%%%%%%%%%%%%%%%%%%%%%%%%%%%%%%%%%%%%%%%%%%%%%%%%%%%%%%%%%%%%

%% file: tables/table2.tex
\begin{table}[!t]
\caption{Ablation studies on different sampling strategies for the auxiliary feature refinement module. For a given value $N$, we first randomly generate $kN$ sampling points on the uncertainty map. Then the top $\beta N$ uncertain points are selected as final sampling points for feature refinement.}
\label{tab1}
\centering
\begin{tabular}{l|c|c|c|c}
\hline
Sampling Strategies & $k$& $\beta$ & Dataset & AP$^{seg}_{\{0.5:0.05:0.95\}}$\\ 
\hline
W.O. Sampling & -- & -- & DSB2018 & 61.14\\ 
Uniform\_small & 1 & 1 & DSB2018 & 61.43\\ 
Biased\_small & 1 & 0.5 & DSB2018 & 60.71\\ 
Uniform & 3 & 3 & DSB2018 & 60.57\\ 
Mildly Biased & 3 & 2 & DSB2018  & 59.85\\ 
Slightly Biased & 3 & 1 & DSB2018 & 60.84\\ 
Moderately Biased & 3 & 0.75 & DSB2018 & 61.25 \\
Greatly Biased & 5 & 0.75 & DSB2018 & 60.94\\
\textbf{Heavily Biased} & \textbf{10} & \textbf{0.75} & \textbf{DSB2018} & \textbf{61.46} \\ 
\hline
W.O. Sampling & -- & -- & Plant & 74.11\\ 
Uniform\_small & 1 & 1 & Plant & 75.40\\ 
Biased\_small & 1 & 0.5 & Plant & 75.58\\ 
Uniform & 3 & 3 & Plant & 75.96 \\ 
Mildly Biased & 3 & 2 & Plant & 76.07\\ 
Slightly Biased & 3 & 1 & Plant & 76.08\\ 
\textbf{Moderately Biased} & \textbf{3} & \textbf{0.75} & \textbf{Plant} & \textbf{76.71} \\
Greatly Biased & 5 & 0.75 & Plant & 76.14\\
Heavily Biased & 10 & 0.75 & Plant &  76.17\\ 
\hline
W.O. Sampling & -- & -- & Neural & 58.26 \\ 
Uniform\_small & 1 & 1 & Neural & 59.26 \\ 
Biased\_small & 1 & 0.5 & Neural & 58.67\\ 
Uniform & 3 & 3 & Neural & 57.81\\ 
Mildly Biased & 3 & 2 & Neural &  59.10\\ 
Slightly Biased & 3 & 1 & Neural & 60.27\\ 
\textbf{Moderately Biased} & \textbf{3} & \textbf{0.75} & \textbf{Neural} & \textbf{60.38}\\
Greatly Biased & 5 & 0.75 & Neural & 59.22\\
Heavily Biased & 10 & 0.75 & Neural & 58.79 \\ 
\hline
\end{tabular}
\label{table2}
\vspace{-1em}
\end{table}

%% file: images/fig7.tex
\begin{figure}[!htb]
\centering
\includegraphics[width=0.8\columnwidth]{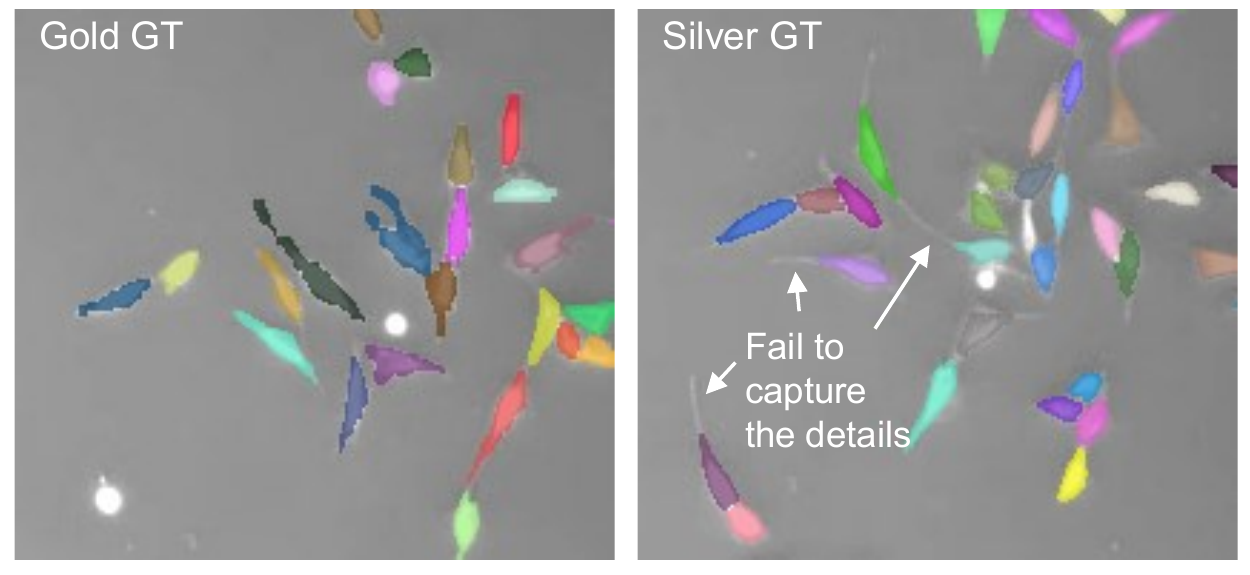}
\caption{Illustration of gold and silver ground-truth annotations for dataset PhC-C2DL-PSC of Cell Tracking Challenge. }
\label{fig7}
\end{figure}

%% file: tables/table3.tex
\begin{table}[!tbh]
        \caption{The performance of our method on the external Cell Segmentation Benchmark of Cell Tracking Challenge. The model is trained separately for different datasets. The A/B format represents our score \cite{RUTG-US} versus the top winner score. GT refers to ground-truth annotations. The silver ground-truth is generated by computer. The gold ground-truth is human-made. The symbol ``\#" refers to the number of images, ``--'' indicates that we do not use the gold annotations, $^*$ represents anonymous contributor.} 
\label{tab1}
\centering
\begin{tabular}{l|c|c|c}
\hline
& Fluo-N2DL-HeLa& PhC-C2DH-U373& PhC-C2DL-PSC\\\hline
Training \# & 184 & 230& 202 \\
Silver GT \# & 184 & 230& 202 \\
Gold GT \# & --&--& 4\\\hline
OP$_{\text{CSB}}$ & 0.924/0.953\cite{HIT-CN} & 0.934/0.959 \cite{CALT-US} & 0.720/0.847$^*$\\%
SEG & 0.863/0.919\cite{HIT-CN} & 0.891/0.927 \cite{CALT-US} & 0.569/0.733$^*$\\%
DET & 0.985/0.992 \cite{KTH-SE} & 0.976/0.991 \cite{CALT-US}  & 0.871/0.972\cite{UVA-NL}\\
\hline
\end{tabular}
\label{table3}
\vspace{-0.5em}
\end{table}

%% file: tables/table4.tex
\begin{table}[!tbh]
        \caption{The performance of our method on the Leaf Segmentation Challenge \cite{CodaLab}. The symbol ``\#'' refers to the number of images. We only use random crop and flipping for data augmentation. We do not use the foreground binary annotations in both training and testing processes.}
\label{tab1}
\centering
\begin{tabular}{c|c|c|c|c|c}
\hline
Datasets & Training \#& bestDice & absDiffFG & diffFG & FgBgDice \\ \hline
A1& 128 &0.88&	1.00&	0.15&	0.96\\
A2& 31& 0.84 &	1.22&	0.11&	0.92\\
A3& 27& 0.81&	0.91&	-0.23&	0.89 \\
A4& 624 &0.86&	0.95&	0.38&	0.94\\
A5& 0 & 0.85&	0.94&	0.20&	0.93\\\hline
Mean & -- & 0.85&	0.95 &	0.21&	0.93 \\
\hline
\end{tabular}
\label{table4}
\vspace{-1em}
\end{table}

%% file: images/fig8.tex
\begin{figure}[!bth]
\centering
\includegraphics[width=\columnwidth]{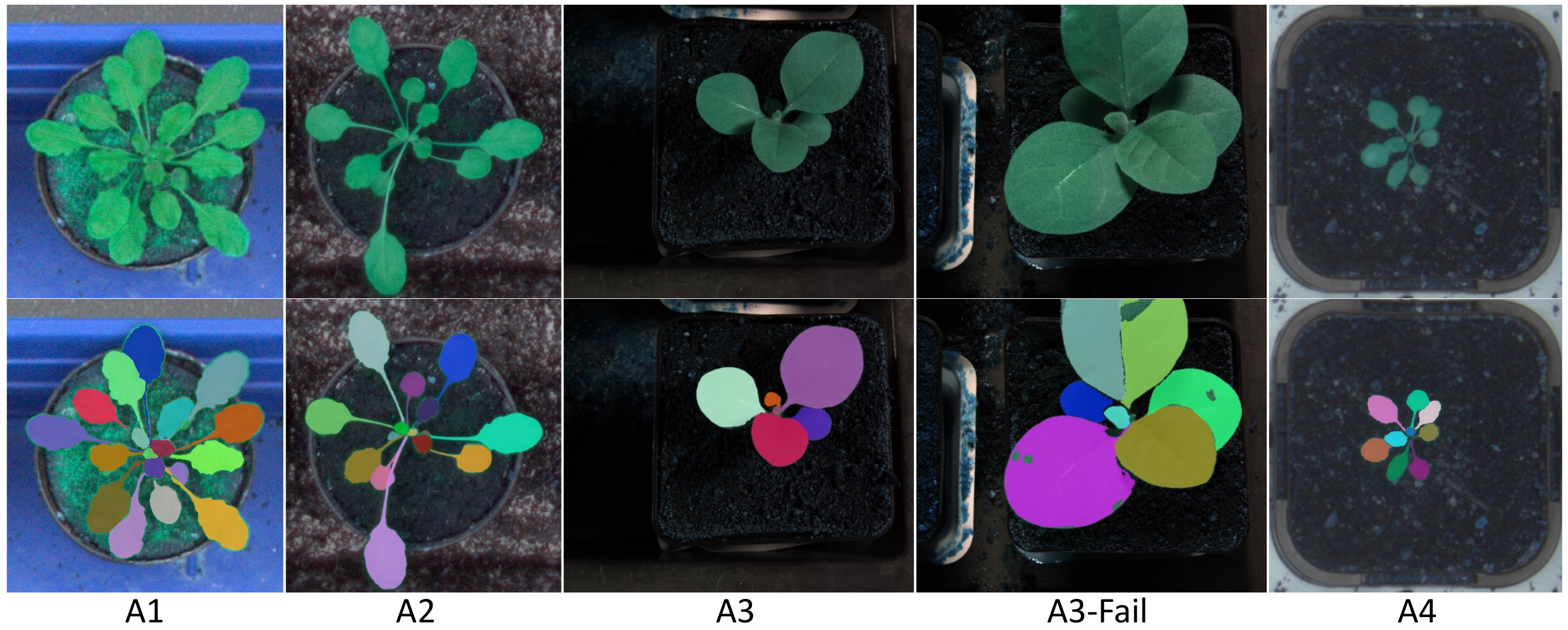}
\caption{Illustration of instance segmentation results on Leaf Segmentation Challenge. We show a failure case for A3 dataset, whose leaves are larger than those in other datasets.} 
\label{fig8}
\vspace{-0.5em}
\end{figure}